\documentclass{article}
\usepackage{graphicx} 
\usepackage{amsfonts}
\usepackage{amsmath}

\title{Large Language Models in Ambulatory Devices for Home Health Diagnostics: A case study of Sickle Cell Anemia Management}
\author{$^1$$^,$$^2$Oluwatosin Ogundare, $^3$Subuola Sofolahan}
\date{%
    $^1$Siemens Technology\\%
    $^2$California State University, San Bernardino\\%
    $^3$Oklahoma State University, Stillwater\\%
    \today
}

\begin{document}

\maketitle

\begin{abstract}

This study investigates the potential of an ambulatory device that incorporates Large Language Models (LLMs) in cadence with other specialized ML models to assess anemia severity in sickle cell patients in real time. The device would rely on sensor data that measures angiogenic material levels to assess anemia severity, providing real-time information to patients and clinicians to reduce the frequency of vaso-occlusive crises because of the early detection of anemia severity, allowing for timely interventions and potentially reducing the likelihood of serious complications. The main challenges in developing such a device are the creation of a reliable non-invasive tool for angiogenic level assessment, a biophysics model and the practical consideration of an LLM communicating with emergency personnel on behalf of an incapacitated patient. A possible system is proposed, and the limitations of this approach are discussed.
\end{abstract}

\section{Introduction}
Every now and again, one might suffer from insomnia, anemia and other symptoms for a variety of reasons. In sickle patients, it might be part of a vaso-occlusive episode. In many cases, the timing of these episodes punctuates the experience especially for acute sufferers. When there is more awareness of the symptoms and what they indicate, a regimen that ameliorates the condition might be easily developed. However, in many developing countries, the living conditions complicates mainstream notions of healthcare regimen and more decisive measures to deliver information are needed. Advancement in nanotechnology and solid state physics have driven the boom in wearable technology or the so-called ambulatory devices. What remains a deterrent for many is that the interpretation of the data that is harvested by the integrated micro and nano sensors are not easily accessible to the layman or intuitive. Many potential customers find a display dominated by numbers without any particular personalized interpretation tedious. In the last decade, fitness and health engineers have begun to incorporate sleep quality assessment into commercial ambulatory devices and in some cases provide information on the onset of a future sleep episode \cite{ogundare2018analysis}\cite{ogundare2019analysis}\cite{ogundare2019statistical} but no real advancement has been made in wearable technology to arrest sudden emergent vaso-occlusive crises in sickle cell patients especially young children in impoverished West African countries. Emergent vaso-occlusive crises often lead to fatalities and is responsible for up to 16\% of mortality in children under the age of 5 years in West African countries especially Nigeria with the documented largest sickle cell gene pool in the world \cite{amoran2017prevention}. Fahlenkamp \& Sofolahan model work on developing a 3D angiogenesis model for biosensing works well in tandem with the prevailing theory that angiogenesis mechanisms are more easily engendered in children with sickle cell anemia as they tend to have elevated angiogenic progenitors \cite{sofolahan2010development} \cite{lopes2015key} \cite{ofori2012elevated}. The ideas in this study are based on the assumption that more angiogenetic activity could impact the echo reflections from acoustic pulses transmitted and received by a nano-engineered acoustic omni-directional transducer and a deep neural network can perform parameter inversion within a reasonable degree of certainty. A discussion of the foundational theory behind our methodology is presented with respect to how Large Language Models (LLMs) improve the state of the art in health monitoring in cadence with other novel applications of LLMs in engineering \cite{ogundare2023industrial}.

\section{Large Language Models and Sickle Cell Management}
We introduce a simple model of the artery without bifurcation as shown in figure 1, within an anatomical region that an extra-corporeal health monitoring device is attached. The length of the referenced anatomical region containing the non-bifurcated artery is $dx$. The arterial walls deform in response to the pressure gradient in the arterial blood vessel such that there is a continuous variation in the arterial radius, $r(x,t)$. The cross-sectional area, $D$, can be computed in terms of the arterial radius as $D(x,t) = \pi(r(x,t))^2$. In a practical computational sense, the length of the reference artery, $dx$, and reference time, $t$, are discretized and spanned by $$i,j \in \mathbb{Z} : i,j =\{1,2,3,\ldots\}$$ such that $$D_{i} = \pi(r(i,j))^2$$

\begin{figure}[ht!]
    \centering
    \includegraphics{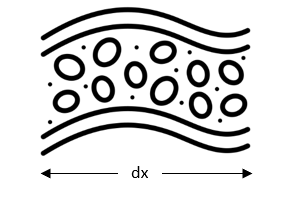}
    \caption{A simple model of non-bifurcated artery in reference region.}
    \label{fig:overview}
\end{figure}

The continuity equation for the reference case, under the assumption that the volume of blood in the artery is fixed within a time slice and an blood flow rate, $u$, will have the following expression for continuity:
\begin{equation}
    \frac{\partial D}{\partial t} +\frac{\partial uD}{\partial x} = 0
\end{equation}
This expression conserves mass if the density, $\rho$, is uniform within the reference arterial length $dx$ but can vary in time such that
\begin{equation}
    \frac{\partial \rho}{\partial t} = 0
\end{equation}
Using Perktold\cite{perktold1991three} rendition of Navier-Stokes equation of viscous flow we have the following momentum equations
\begin{equation}
    \frac{\alpha^2}{Re}\frac{\partial \textbf{u} }{\partial t} + (\textbf{u} \cdot \nabla)\textbf{u} + \nabla \textbf{p} - \frac{1}{Re} \Delta \textbf{u} = 0
\end{equation}
Coupled with a ZERO divergence of the velocity field. $\nabla \textbf{p}$ is the pressure gradient and $Re$ is the Reynold's number and $\alpha$ is the Womersley number.\\ \\
The main idea presented in this section is that an acoustic source can create a transient pressure wave whose reflections in the reference arterial region is a function of the arterial radii field, $r(x,t)$. We claim that 
\begin{equation}
 differential \hspace{1mm} pressure \hspace{1mm} gradient \hspace{1mm} field, P(x, r, t) = f(r(x,t))
\end{equation}
Consequently, the classical solution in the case of a single harmonic frequency, $\omega$ is given by
\begin{equation}
P(x,r,t)=Ae^{(i(\omega t- k_x x-k_r r ))}+ Be^{(i(\omega t- k_x x + k_r r ))}
\end{equation}
Such that $Ae^{(i(\omega t- k_x x-k_r r ))}$ is the wave travelling in the +x direction (frontier wave) and $Be^{(i(\omega t- k_x x + k_r r ))}$ is the wave travelling in the -x direction (reflected wave) (see Figure 2 below) and both are solutions to the classical wave equation and $k_x$ and $k_r$ = wave number. \\ \\

\begin{figure}[ht!]
    \centering
    \includegraphics{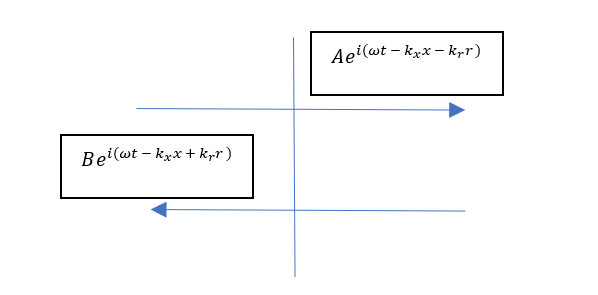}
    \caption{Interface showing Forward Propagating Wave and Reflected Wave.}
    \label{fig:overview}
\end{figure}

For practical consideration, we simply take $A$ and $B$ to be the mean power of the generated incident pulse from the nano-engineered pulse generator shown in figure 3. The observed echo reflections from the incident pulse are recorded and inverted to determine the set of related arterial radius using a Deep Physics neural network model that is trained to solve the inverse problem using the values of the observed pressure field. \\ \\
Acoustic "Time of flight" measurements are recorded with every reading and used to determine changes in blood density as a possible indicator of increased angiogenesis progenitors. The data from the Deep Physics model estimating the variation of the radii field due to the pressure gradient in the arterial blood vessel as well as fractional changes in the blood density between measurements is fed to a Large Language Model (LLM) that estimates the likelihood of a vaso-occlusive episode over a pre-defined time scale.\\ \\
Mathematically, Let $V$ be such that\\

\begin{equation}
   V =\begin{cases}
			1, & \text{Vaso-occlusive Episode}\\
            0, & \text{otherwise}
		 \end{cases}
\end{equation}
.\\ \\ \\ \\

\begin{figure}[ht!]
    \centering
    \includegraphics[width=\textwidth]{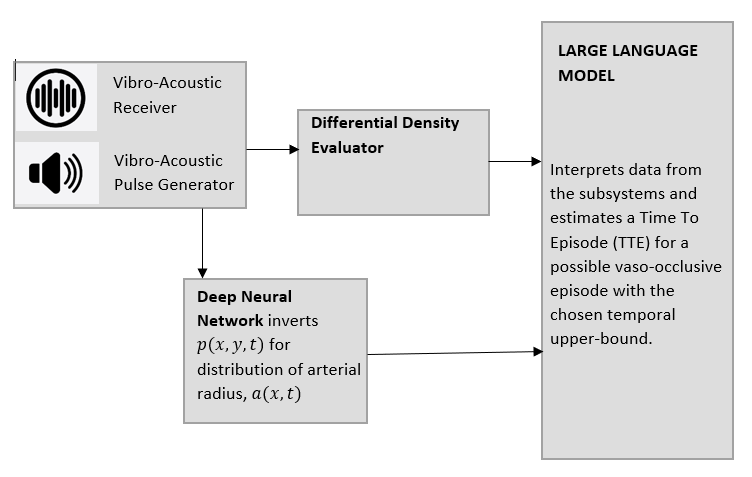}
    \caption{A schematic of the Adaptive Sickle Cell Monitor.}
    \label{fig:overview}
\end{figure}

The LLM estimates the likelihood that at time, $t = 0$ (present) the patient is having a vaso-occlusive episode, i.e. $Pr(V = 1 \hspace{1mm}|\hspace{1mm} t = 0, Biophysics\hspace{1mm}Data)$ and compares it with the likelihood of having the episode at regular intervals in the future with $t$ taking on discrete values from $\mathbb{Z^+}$ i.e., \{$1, 2, 3, \ldots$\}. The chosen time period is the one that maximizes the probability. An upper bound is chosen to limit how far in the future the algorithm is needed.
\begin{equation}
   Time \hspace{1mm} To \hspace{1mm} Episode(TTE) = \max_{i \in \{1, 2, 3, \ldots\}} Pr(V = 1 \hspace{1mm}|\hspace{1mm} t = i, Biophysics\hspace{1mm}Data)
\end{equation}
In practice the actual temporal gap between the discrete time steps can be tuned as needed. In addition to the LLM computing the TTE, it is calibrated to provide informal recommendations on the next steps based on the estimated TTE. Where the infrastructure allows, the LLM has the option to contact the emergency hotline and communicate with emergency personnel on behalf of a patient.

\section{Conclusion}
Overall, the proposed ambulatory device has the potential to significantly improve sickle cell management by providing real-time information on anemia severity and reducing the frequency of vaso-occlusive crises by incorporating Large Language Models (LLMs) and specialized ML models to assess anemia severity in sickle cell patients in real-time. Reliable nano-engineered sensors are critical to dependably measure indicators of elevated angiogenesis progenitors levels to assess anemia severity and to provide real-time information to patients and clinicians to reduce the frequency of vaso-occlusive crises by allowing for timely interventions and potentially reducing the likelihood of serious complications. While the development of such a device presents some challenges, especially the creation of a reliable nano-engineered vibro-acoustic sensing device, a biophysics model, and the practical consideration of an LLM communicating with emergency personnel on behalf of an incapacitated patient, the potential benefits of such a device for sickle cell management make it a worthwhile endeavor. With further research and development, this technology has the potential to improve the lives of many sickle cell patients in developing countries.

\bibliographystyle{plain} 
\bibliography{refs} 
\end{document}